\documentclass[10pt,journal,compsoc]{IEEEtran}
\ifCLASSOPTIONcompsoc
  \usepackage[nocompress]{cite}
\else
  \usepackage{cite}
}
\fi
\ifCLASSINFOpdf
  \usepackage[pdftex]{graphicx}
\else
\fi
\usepackage{url}

\usepackage{amsmath}

\begin{document}
%
\title{Machine Learning Capability: A standardized metric using case difficulty with applications to individualized deployment of supervised machine learning}
%
%
%
%

\author{Adrienne~Kline*,~\IEEEmembership{Member,~IEEE,}
        and~Joon~Lee,~\IEEEmembership{Senior Member,~IEEE}
\IEEEcompsocitemizethanks{\IEEEcompsocthanksitem A. Kline is with the Department of Biomedical Engineering, Data Intelligence for Health Lab, Cumming School of Medicine, University of Calgary, Calgary, AB, Canada
\protect\\
corresponding author* e-mail: askline1@gmail.com
\IEEEcompsocthanksitem J. Lee is with the Data Intelligence for Health Lab, Department of Community Health Sciences, and Department of Cardiac Sciences, Cumming School of Medicine, University of Calgary, Calgary, AB, Canada}
\thanks{}}

\IEEEtitleabstractindextext{%
\begin{abstract}
Model evaluation is a critical component in supervised machine learning classification analyses. Traditional metrics do not currently incorporate case difficulty. This renders the classification results unbenchmarked for generalization. Item Response Theory (IRT) and Computer Adaptive Testing (CAT) with machine learning can benchmark datasets independent of the end-classification results. This provides high levels of case-level information regarding evaluation utility. To showcase, two datasets were used: 1) health-related and 2) physical science. For the health dataset a two-parameter IRT model, and for the physical science dataset a polytonomous IRT model was used to analyze predictive features and place each case on a difficulty continuum. A CAT approach was used to ascertain the algorithms’ performance and applicability to new data. This method provides an efficient way to benchmark data, using only a fraction of the dataset (less than 1\%) and 22-60x more computationally efficient than traditional metrics. This novel metric, termed Machine Learning Capability (MLC) has additional benefits as it is unbiased to outcome classification and a standardized way to make model comparisons within and across datasets. MLC provides a metric on the limitation of supervised machine learning algorithms. In situations where the algorithm falls short, other input(s) are required for decision-making. 
\end{abstract}

\begin{IEEEkeywords}
Item Response Theory, Machine Learning, Model Evaluation, Computer Adaptive Testing, individualized machine learning, ethical artificial intelligence
\end{IEEEkeywords}}

\maketitle

\IEEEdisplaynontitleabstractindextext

%
\IEEEpeerreviewmaketitle

\IEEEraisesectionheading{\section{Introduction}\label{sec:introduction}}

\IEEEPARstart{C}{lassical} test theory (CTT) is the current modus operandi for evaluating supervised machine learning models, that assumes all cases (samples) are created equivalent. This approach is operationalized by first establishing a percentage of the data held in reserve (i.e., ‘test data’) to which the model is naive. The test data are comprised of a random sample of anywhere between 10-30\% of the total dataset and capitalizes on an n-fold cross validation process where the portion of data allocated for evaluation purposes is staged and iterated upon \cite{Krstajic}. From this point evaluations are made to assess model performance. These typically include area under the curve (AUC), receiver operating curve (ROC), accuracy, sensitivity, specificity, recall, and F1. 
While these evaluation metrics are useful, they are not without their drawbacks \cite{Santafe}. These processes suffer from two pitfalls: 1) it requires the use of all the data in the set due to concerns regarding bias and undergoes cross-validation to compensate and 2) it does not provide an understanding of how robust the model is in the face of a case of varying classification difficulty.\\

Item response theory (IRT) is synonymous with modern test theory. Within this field questions administered to students on exams are scored either 'correct' or 'incorrect', binarizing the data. The pattern of responses for any given question are used to develop logistic regression curves that delineate the difficulty of the question posed and subsequently the latent ability of an examinee for a given question \cite{Embretson_2000}. IRT models based on 0 or 1 outcome data are referred to as dichotomous IRT models which include one-, two- and three-parameter models. The one-parameter logistic model (1PL) has only one characteristic often referred to as the difficulty, denoted with \textit{b} \cite{Rasch_1960}. The b-characteristic is scaled using the normal distribution. Questions (items) with higher \textit{b} values require more information or a higher latent ability to get correct and score '1'. Contrast this with polytonomous IRT models \cite{Samejima_1969} which allow for non-binary outcomes referred to as a graded response models where false dichotomies are generated that collapse feature options (1, 2, 3, 4 for a 4 outcome variable which can be continuous or categorical) into two groups at various stages along the value continuum. Both dichotomous and polytonomous IRT models can be extended into the machine learning literature where we can think of individual cases as students taking exams and liken the predictor variables to the questions posed creating cut points for right and wrong and will be showcased here.\\

The purpose of this study was to capitalize on integrating IRT and Computer Adaptive Testing (CAT) with machine learning. IRT provides estimates of case difficulty. CAT is normally used to present items of varying difficulty to test takers, and programs have been written to automate the CAT process \cite{Mengehetti}; however, we presented cases of varying difficulty, in terms of their ease of classification determined a-priori and classification independent, within a machine learning environment. We compared the results of a traditional machine learning training/testing approach with our approach.\\

IRT has been used in machine learning environments to select features that have characteristics contributing to classifying cases \cite{Kline_features}. It has also been shown to be useful in identifying case classification difficulty \cite{Kline_JMIR}. Case difficulty refers to the level of difficulty a supervised machine learning classification model will have in correctly classifying the case. The focus of the current research is to extend the use of the case classification difficulty parameters that are estimated as part of the IRT analyses. While these difficulty parameters are often called person-theta values ($\theta$) in traditional IRT contexts, in the current study we will refer to them as Case Difficulty Indices (CDIs). CDI values for a sample of cases that have been IRT analyzed based on a number of features are normally distributed with a mean of 0 and standard deviation of 1.0. Thus, most CDIs range between -2.50 and +2.50. 
The CDI results allow for the identification of which cases are more centrally and more peripherally located on the difficulty distribution. Assuming a binary classification model, cases that are more centrally located are more difficult to classify into their respective categories, while those at the periphery are easier \cite{Kline_JMIR}. This phenomenon is shown in Fig. 1, where the probability of correct classification for those cases who are in class 2, gets progressively better as estimated CDI values become higher, and progressively worse as estimated CDI values become lower. Conversely, for those cases in class 1, the probability of correct classification gets progressively worse as CDI estimates become higher, and progressively better as estimated CDI values become lower.\\
\begin{figure}[!h]
\centering
\includegraphics[width=7cm]{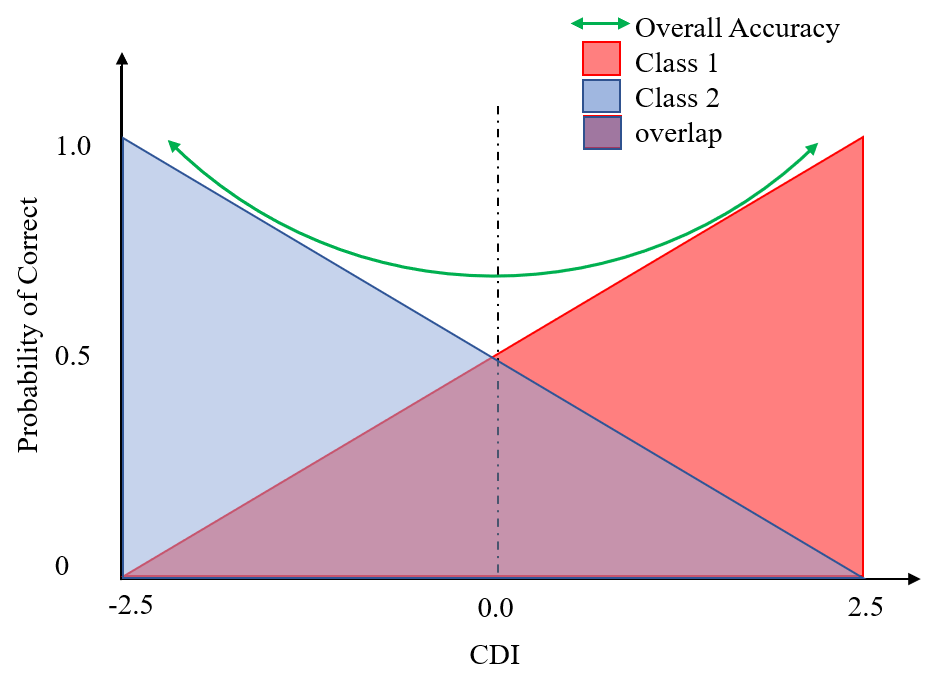}
\caption{Probability of correct classification vs. CDI score for each class}
\end{figure}

Because the CDI estimates are based on the normal distribution, scores below the mean of 0.0 have negative (low) values and those above the mean have positive (higher) values. Thus, one of the classes of cases must have their CDI values multiplied by -1 (to obtain the additive inverse of the values) to assure comparable interpretation for both classes when graphing the data and later for administration of a CAT.\\

Also capitalizing on the a-priori case difficulty information, we can take advantage of an adaptive approach to machine learning model evaluation. That is, rather than allowing the machine learning model to randomly select cases as input for classification testing, we take a more theory-driven approach and guide the machine to self-select from cases at specified CDI levels. This novel approach will provide a unique metric, the Machine Learning Capability (MLC), by which to evaluate the utility of the machine learning model and information about the data set itself.\\
To execute this process, we draw on the principles of Computer Adaptive Testing (CAT) for guidance, as this is a long-standing methodology developed for ascertaining a test-taker's innate ability. CAT was developed to provide a precise and efficient estimate of a test-taker’s ability \cite{Lord}. CAT begins with the creation of a population of items with known characteristics (such as level of difficulty). The principle of applying CAT works as follows: 1) a question of intermediate difficulty is offered first to the test taker; 2) the test taker gets the item correct or incorrect; 3) If an incorrect response is made, the next item presented will be easier; 4) If a correct response is made, the next item presented will be more difficult; 5) This process continues iteratively, initially making larger steps between levels of difficulty and then smaller and smaller steps between item difficulty levels; and 6) Stopping criteria are invoked to end the process.\\
An overview of CAT Methodology is shown in Fig. 2. An initial estimation of the latent ability of the test taker is provided to start the process. When administering the CAT in machine learning, there is an expected monotonic progression in the probability of incorrectly classifying the case as CDI values increase \cite{Kingsbury}.

\begin{figure}[ht]
\centering
\includegraphics[width=8cm]{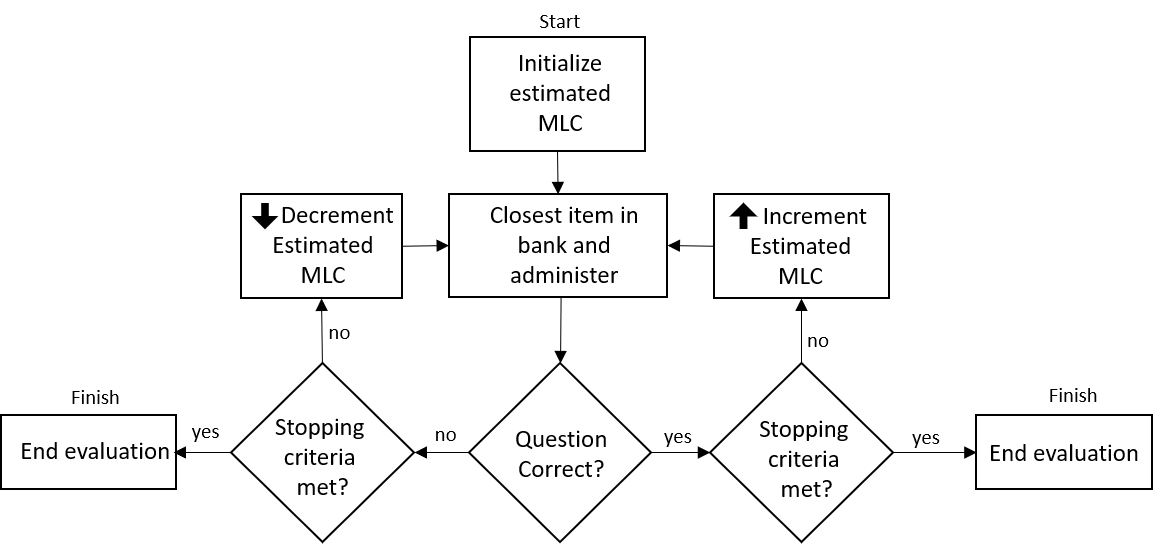}
\caption{Overview of Computer Adaptive Testing (CAT) Methodology}
\end{figure}

We employed a CAT-based evaluation of machine learning models by using cases, rather than test items, as the input. In particular, the relative difficulty levels of the cases and how the machine learning models performed on them was the focus of this research. 
The model evaluation criteria included time allocated to the evaluation, the number of cases needed to ascertain an estimate of the model’s classification performance, the MLC metric that incorporates how robust the model is when presented with easy and difficult cases. \\

\section{Methods}
We deliberately used two quite different datasets to demonstrate this process. One was a health-related data set, the Medical Information Mart for Intensive Care III (MIMIC-III). This set contains numerous features modestly related to the class outcome (in-hospital mortality). The other was a physical science data, the Predicting Pulsar Star data set obtained from the UCI Machine Learning Repository. This set has only a few features but they are strongly related to the class outcome. Doing so showcased how such differences play out in machine learning models but are not explicit using traditional evaluation metrics. \\
We also deliberately created balanced data sets. This is because we wanted to ascertain the intrinsic ability of the classifier with respect to class without biasing it. \\

\subsection{MIMIC Data Set}
The MIMIC-III database houses health data of over 40,000 critical care unit patients at the Beth Israel Deaconess Medical Centre between 2001-2012 \cite{Johnson} \cite{Johnson_2017}. The database was queried using SQL plug-in for Python. Case inclusion criteria were: 1) age of 16 years or older; and 2) and at least 3/4 of the features of interest available for a select case (patient), leading to subsequent imputation. There were 4039 cases over the age of 16 that experienced ‘death in hospital.’ From the remaining ‘no death in hospital’ data a random sample was selected equivalent to the number that died to give a final balanced sample size of 8,078. This data set was used previously \cite{Kline_JMIR} to demonstrate how classifiers deal with different levels of cased difficulty. \\

\subsubsection{MIMIC Features}
Features of predictive interest for the machine learning model were selected based on the SAPS II (simplified acute physiology) score \cite{LeGall} and are shown in Table 1. They included demographic, procedural, pre-existing conditions, and laboratory values. Laboratory values represent the worst-values taken during the ICU stay. \\
To prepare the data for a dichotomous IRT analysis, all features were coded into disease promoting (1) or disease protective (0) states. Lab values falling outside of the normal ranges were coded with a 1 (too low or too high), disease presence (e.g. Metastatic cancer) was coded with a 1, age was demarcated based on those above 65 who were allocated with 1 \cite{Fuchs}\cite{Lindemark}\cite{Alam}\cite{Atramont}, men were assigned 1 \cite{Mahmood}, and those that entered under emergency were coded 1. Normal values were obtained from several sources \cite{MedicalCouncil}\cite{Teasdale_1974}\cite{Teasdale_1976}\cite{Ichan} \cite{Lapum}\cite{MDCalc}\cite{Healthline}.

\begin{table}[ht]
\caption{MIMIC Variables - Based on SAPSII}
\label{table_example}
\centering
\begin{tabular}{|c||c|}
\hline
Feature & Healthy normal = 0\\
\hline
AIDS & Absent = 0, Present = 1\\
Heme Malignancy & Absent = 0, Present = 1\\
Metastatic Cancer & Absent = 0, Present = 1\\
Glasgow coma scale (1-15) & 15 = 0, $\leq$ 14 = 1\\
WBC Minimum ($*10^9$/L) & 4-10 = 0\\
WBC Maximum ($*10^9$/L) & 4-10 = 0\\
Na Minimum (mmol/L)& 135-145 = 0\\
Na Maximum (mmol/L)& 135-145 = 0\\
K Minimum (mmol/L)& 3.5-5 = 0\\
K Maximum (mmol/L) & 3.5-5 = 0\\
Bili Maximum (mg/dL)& $\leq$ 1.52 = 0\\
HCO$_3$ Minimum (mmol/L) & 24-30 = 0\\
HCO$_3$ Maximum (mmol/L)& 24-30 = 0\\
BUN Minimum (mg/dL)& 7-22 = 0\\
BUN Maximum (mg/dL)& 7-22 = 0\\
PO$_2$ (mmHg)& 85-105 = 0\\
FiO$_2$ (\%) & 21 = 0, $\leq$ 21 = 1\\
Mean heart rate (bpm) & 60-100 = 0\\
Mean systolic blood pressure (mmHg) & 95-145 = 0\\
Maximum temperature ($^{\circ}$C) & 36.5-37.5 = 0\\
Urine Output (mL/24h)& 800-2000 = 0\\
Sex (M/F) & M=1, F=0\\
Age (yrs) & $\leq$ 65 = 0, $>$ 65 = 1\\
Admission type & Emergency = 1, else = 0 \\

\hline
\end{tabular}
\end{table}

\subsection{Pulsar Data}
Pulsars are a rare type of Neutron star that produce detectable radio emissions. They are of scientific interest as probes of space-time, the inter-stellar medium, and states of matter. Because of the rarity of pulsar stars, detecting the broadband radio emission signature generated by a true pulsar from other-generated broadband signatures is an important classification task.\\
The Predicting Pulsar Star data set (N=17898), obtained from the UCI Machine Learning Repository (N=17898) \cite{Dua}\cite{Lyon_2017}\cite{Lyon_2016} was used in this study. Of the total cases, 1639 (9.2\%) were pulsars and 16259 (90.8\%) were not. A balanced data set was created by selecting a random sample of 1639 cases from the non-pulsar stars, providing a final sample size of 3278.\\

\subsubsection{Pulsar Features}
Four features were used from the data set. They are continuous variables obtained from the integrated pulse profile that describe a longitude-resolved version of the signal averaged in both time and frequency. They include the Mean, Standard Deviation, Excess Kurtosis, and Skewness of the integrated profile. Because of the relationships between the features and classification, the mean and standard deviation scores were inverted (multiplied by -1) \cite{Mir}, rendering all features monotonically and positively related to the classification categories of pulsar =1/non-pulsar = 0. To prepare the data for a polytomous IRT analysis, the continuous data of the four predictors were divided into their respective quartiles coinciding with 0-25\%, 25-50\%, 50-75\% and 75-100\% (Table 2).

\begin{table*}[ht]
\caption{Pulsar Feature Cutpoints for the Integrated Profile}
\centering
\begin{tabular}{|c||c|c|c|c|}
\hline
Quartile & Mean & Standard Deviation & Kurtosis & Skew\\
\hline
0-25\%  (1)	& Lowest to -118.42383 &	Lowest to -49.051276 &	Lowest to .176350204 &	Lowest to .082984570\\
26-50\% (2) &	-118.42384 to -95.382813 &	-49.051277 to -43.567262 &	.176350205 to .633206915 &	.082984571 to 1.31210544\\
51-75\% (3) &	-95.382814 to -54.195313 &	-43.567263 to -36.387227 &	.633206916 to 2.96592054 &	1.31210545 to 11.6310510\\
76-100\% (4) &	-54.195314 to Highest &	-36.387228 to Highest &	2.96592055 to Highest	& 11.6310511 to Highest\\
\hline
\end{tabular}
\end{table*}

\subsection{IRT Analyses}
IRTPRO\cite{Cai} was used to obtain the difficulty and discrimination characteristics of the predictor variables which subsequently underpins the scoring system of case difficulties in each of the data sets. These calculated predictor variable characteristics combined with each case’s scores on the features provide a pattern of responding that allows for unique overall scoring for each case.\\

\subsubsection{IRT Feature Analysis}
The MIMIC predictor data were encoded to binary (0,1) in accordance with the normal value cut points mentioned previously. Thus, a 2-parameter logistic marginal maximum likelihood model was used to determine the difficulty ($\beta_i$) and discrimination parameters ($\alpha_i$)\cite{Bock_1981} for each predictor variable/item. This is shown in the logistic curve as per Eq. 1. $X_{is}$ is the response of the person/case instance ($s$) to the specific item/predictive variable '$i$' (0 or 1), $\beta_i$ the difficulty of item $i$ (location of the inflection on the x-axis) and $\alpha_i$ the discrimination or slope of the item. This is visually demonstrated in Figure 3. 

\begin{figure}[ht]
\centering
\includegraphics[width=7cm]{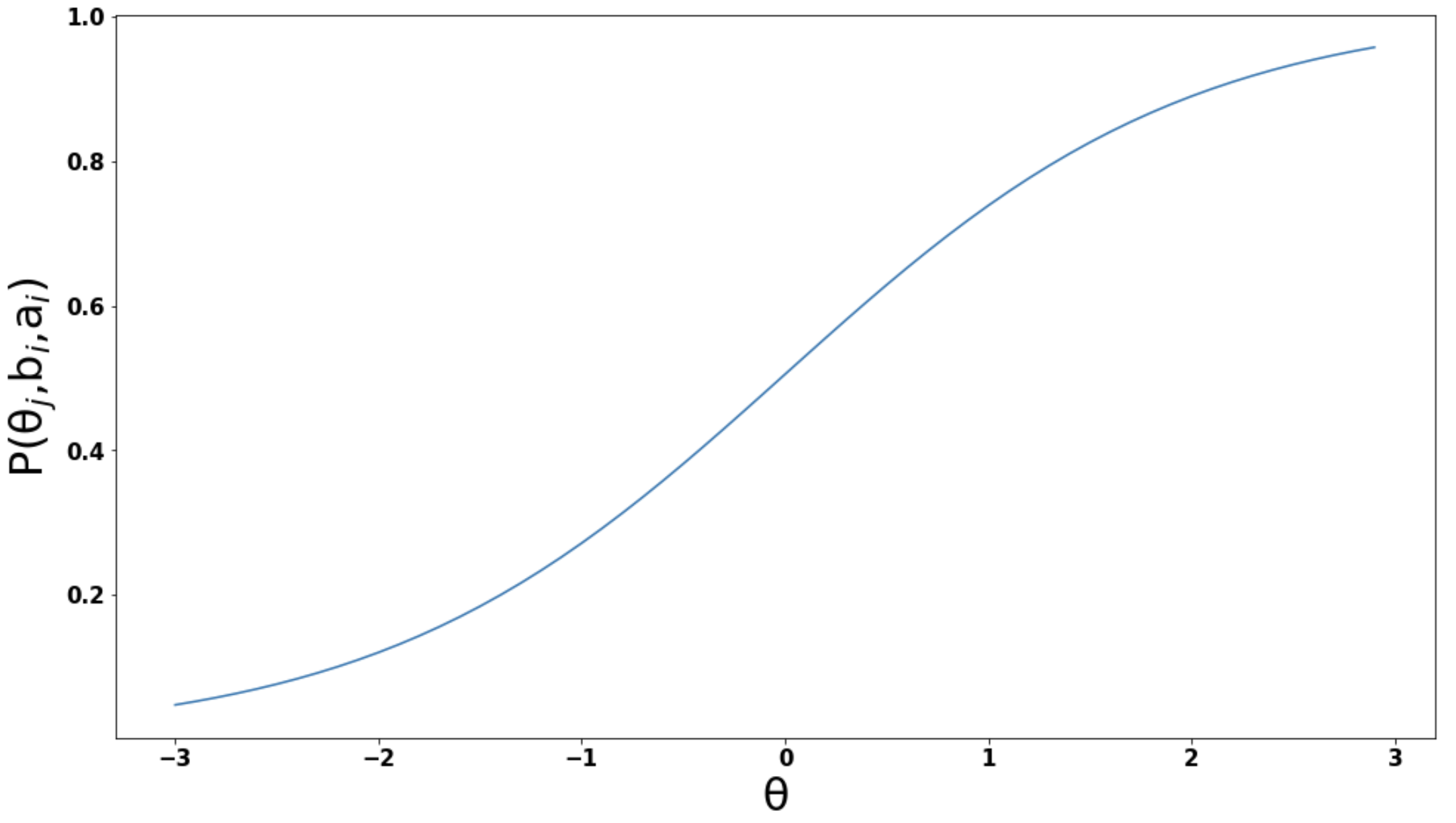}
\caption{Example of an item characteristic curve in 2-PL model}
\end{figure}

\begin{equation}
	P(X_{is}=1|\theta_s,\beta_i,\alpha_i) = \frac{\exp[\alpha_i(\theta_s - \beta_i)]}{1 + \exp[\alpha_i(\theta_s - \beta_i)]} 
\end{equation}

The Pulsar data were continuous and separated into 4 quartile based categories (0,1,2,3) as this set of predictor variables can be characterized as ordered categorical responses. Thus, a polytomous maximum likelihood graded response model \cite{Samejima_1969}, \cite{Samejima_1972}, \cite{Penfield} was used to determine the difficulty and discrimination parameters of the predictors for each of the 4 categories. In a graded response model, each item/predictive feature is described by a common item slope parameter ($\alpha_i$) and one between category threshold $\beta_{ij}$) where $j=1,..m_i$, and $m_i+1$ is equal to the number of response categories within an item/predictor, where each category can now be resembled by a 2-PL model. Computing the conditional probability for each case responding to a particular category requires a two-step process. 1) Computation of ($m_i$) distinct curves for each item (Eq. 2), and 2) computing the actual category response probabilities for $x = 0,1,2,3$ by subtraction of all categories of greater value (Eq. 3). 

\begin{equation}
	P_{ix}^*(\theta_s) = \frac{\exp[\alpha_i(\theta_s - \beta_ij)]}{1 + \exp[\alpha_i(\theta_s - \beta_ij)]} 
\end{equation}
\begin{equation}
	P_{ix}^*(\theta_s) = P_{ix}^*(\theta_s) - P_{i(x+1)}^*(\theta_s)
\end{equation}

\begin{figure}[ht]
\centering
\includegraphics[width=7cm]{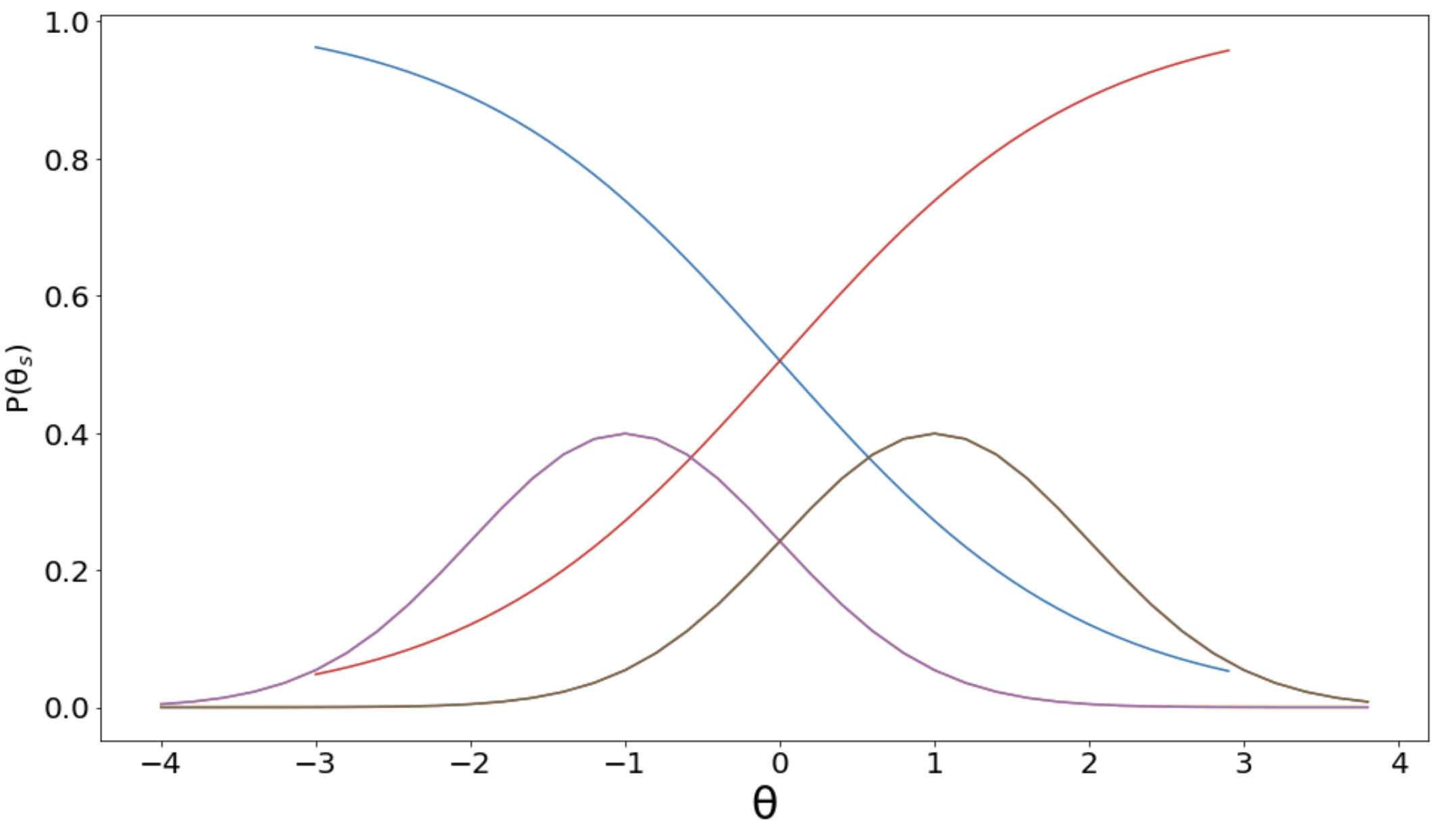}
\caption{Example of an item characteristic curves in polytonomous model}
\end{figure}

\subsubsection{IRT Case Difficulty Index (CDI) Analysis}
Case difficulty indices (CDI) are synonymous with the latent case based ability (referred to as $\theta$ throughout these equations. This distinction is required given the classification nature of a machine learning problem and how it differs from psychometric testing usage. Additionally it has been cited in the literature previously \cite{Kline_JMIR}.\\
CDIs are calculated using a maximum likelihood estimation. Each case provides 'responses' to questions (predictor variables) with each response being either correct or incorrect (as converted previously based on normal values or scale). Therefore, the probability that a case with CDI ($\theta$) level obtains a given response on that variable can be computed. The probability that a case with CDI ($\theta$) obtains a response $U_i$ on item $i$ where $U_i$ = 0 if incorrect and $U_i$ = 1 if correct is denoted by $P(U_i | \theta)$. In a correct response ie. $P(U_i=1| \theta)$ this can be denoted $P_i(\theta)$, and as $U_i$ is a binomial variable this relationship can be expressed as:

\begin{equation}
   P(U_i | \theta) = P_i^{U_i}(1-P_i)^{1-U_i}
\end{equation}
\begin{equation}
   P(U_i | \theta) = P_i^{U_i}Q_i^{1-U_i}
\end{equation}
The probability of the vector of responses provided can be given by Eq 6, and taking the natural log results in equation 7.
\begin{equation}
   P(U_i | \theta) = \prod_{i=1}^{n}P_i^{u_i}Q_i^{1-u_i}
\end{equation}
\begin{equation}
   L = log(P(U_i | \theta))
\end{equation}

The value of $\theta$ that maximizes the likelihood function L is $\hat{\theta}$ or latent ability/CDI:
\begin{equation}
   \theta= \underset{\theta}{argmax}(L)
\end{equation}

It is the response pattern of a case to the items (predictor variables) parameters $\alpha$ and $\beta$ that ultimately offer placement of the case along the CDI continuum (alive-dead or non-pulsar-pulsar). Lower CDIs indicate the case is less likely to be either a dead case for the MIMIC data or a pulsar star for the Pulsar data. Conversely, higher CDIs indicate the case is more likely to be either a dead case for the MIMIC data or a pulsar star for the Pulsar data. As noted earlier, CDI scores typically range from -2.5 to + 2.5.\\

\subsection{Machine Learning, Computer Adaptive Testing and the Machine Learning Capability (MLC) Metric}
To compare the traditional machine learning evaluation paradigm to the one proposed in this research, the data were processed in two different ways.

\subsubsection{Traditional Machine Learning Analysis and Evaluation}
Machine learning and all subsequent analyses were performed in a Python 3.7 environment. Using a traditional approach, both datasets underwent training on a neural network. To obtain the traditional evaluation metrics we split the data in a random 70-30 train-test split for use in ascertaining hyperparameters, where 70\% was used in development. Hyperparameters were selected using a grid search for learning rate, activation function and number of neurons in the hidden layer. The grid investigated for the neural network included activation functions such as softmax, softplus, softsign, relu, Tanh, sigmoid, and hard sigmoid; learning rates such as 0.01, 0.1, 0.2, and 0.3; and hidden neurons 
of 6, 8, 12, 18, 24, and 30. Within this hyperperparameter tuning approach, the 70\% dataset underwent a 5 fold cross-validation and model evaluation was performed using the traditional metrics AUC, recall, accuracy and F1. Following this the same metrics were calculated based on the 30\% hold out model naive test set.\\

\subsubsection{Machine Learning Analysis Evaluated Via CAT}
The CDI values from each data set were placed on a their respective continua. Then CDI scores were ‘binned’ in increments of 0.25$\sigma$ intervals centered at zero. Cases in each of the bins were randomly split into 70-30 (train-test). This was done to assure that there was a stratified random sampling based on case difficulty used to train the model. Bins with a very small number of cases ($\leq$2) were either split in half for training/testing if there were only two cases in the bin or assigned to the training set if there was only one case in the bin. The same hyperparameters as those from the traditional approach were used to ascertain the hyperparameters of this model. This model did not use the 30\% test sample to generate traditional metrics. Instead, once the neural network model was trained on the 70\% training cases, the CAT procedure was initiated.\\ 
Thus, the CAT evaluations were based on the 30\% test cases that had been ‘binned’. This was done so that is was possible to present increasingly difficult or easier cases to the model to which the model was naive. Two sets of CAT analyses were conducted for each of the two data sets (MIMIC and Pulsar), as each class of data was run separately within the CAT framework. Class 1 included the ‘dead’ cases in the MIMIC and ‘pulsar’ cases in the Pulsar data sets. Class 2 included the ‘alive’ cases in the MIMIC and ‘non-pulsar’ cases in the Pulsar data set. Each of these four classes underwent separate CAT evaluation protocols.

One of the classes must have their CDI values multiplied by -1 (to obtain the additive inverse of the values) to assure comparable interpretation for both classes administrating the CAT. This ensures a unidirectional representation of difficulty. Thus, the class 1 data (‘dead’ in the MIMIC dataset and ‘pulsar star’ in the Pulsar dataset) were multiplied by -1 to ensure comparable CDI values across class. Figure 3 shows the inversion of scores from Fig. 1 for the class 1 CDI scores. \\

\begin{figure}[ht]
\centering
\includegraphics[width=7cm]{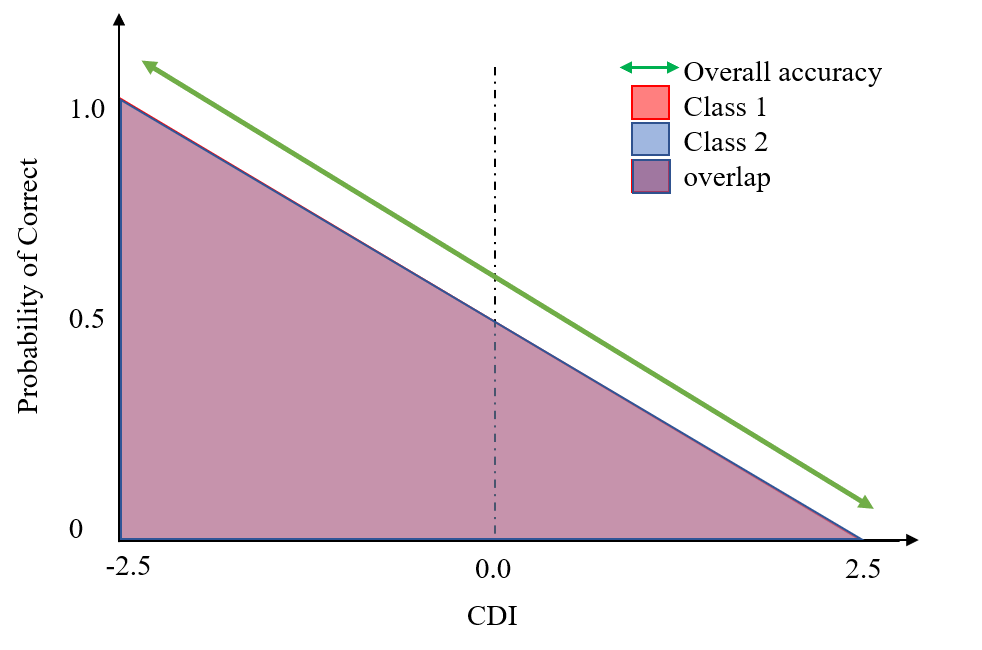}
\caption{Probability of correct classification vs. CDI score for each class}
\end{figure}

We then presented a case from each class with CDI values closest to their 25th percentile to initialize the algorithm for each class. Cases from each class were then presented for classification to the algorithm one at a time, based on their CDI value, which was dependent on whether the algorithm correctly or incorrectly classified the prior case. The evaluation process stored the: 1) total number of cases queried in increments of 1 after each case is 'presented for classification' by the algorithm, 2) number of correct and incorrect classifications and 3) cumulative sum of the CDIs presented. \\
Case presentation was performed in a step-wise fashion \cite{Wright} using a simple algorithm particularly useful in a Rasch IRT model \cite{Mead} - which characterizes our case-difficulty data - and the purpose is to classify \cite{Linacre}. In the event the case was correctly classified, Eq. 6 is used. If it is incorrectly classified equation Eq. 7 is used, where L denotes the number of cases presented to that point, and G is the addition of Gaussian randomness inserted into the increment but whose standard deviation is within the prescribed stopping criteria estimate. We added this to introduce some randomness into the next estimate of ability to avoid a local minimum.

\begin{equation}
    CDI_{t+1} = CDI_t + \frac{2}{2^L} + N(0,0.1)
\end{equation}

\begin{equation}
    CDI_{t+1} = CDI_t - \frac{2}{2^L} + N(0,0.1)
\end{equation}

The criterion for when the testing process ends is synonymous with the stopping rule. Typically, this is either after a certain number of test items have been administered (fixed-length) or after a specified precision of the estimator has been reached (variable length). A variable length stopping criteria was used (standard error of the measure), seen in Eq. 8 below. Where $SE_M$ denotes the standard error of the measure, and $r$ is the reliability of the measure that is set by the user at outset. For our purposes $\sigma$ = 1, as the CDI values have already been normalized ($\mu$ = 0; $\sigma$ = 1) and it was determined that we wanted a reliability of 0.98 for our $SE_M$. The smaller the $SE_M$ the more cases it will take to solve, but the more reliable the result. Utilizing these restrictions the $SE_M$ was determined to be 0.14. The standard error of the measure was calculated on each iteration and if less than the value set by the user, the process will terminate.

\begin{equation}
	SE_M = \sigma\sqrt{1-r}   
\end{equation}

We have termed the metric associated with the capability of the supervised algorithm as the Machine Learning Capability (MLC). Recall that the CAT ends when the highly reliable highest difficulty is reached. Thus, this new metric indicates the case difficulty level at which the machine learning model terminates. It is calculated for each class, offering the user an understanding of the capability of the model for each. Computational efficiency was calculated as the total time it took for each case to be processed through the network.\\
Estimation of the MLCs for each class is calculated by Eq. 9. The code was written so it only becomes eligible to be calculated after a minimum of five cases have been used and is updated iteratively in a loop. Modified versions of Eq. 9 were used if the total number of right or wrong cases were equal to zero (Eq. 10 and Eq. 11, respectively). Where $H$ is the summed difficulty of cases asked, $L$ the number of cases used, $R$ the number of cases correctly classified and $W$ the number of cases incorrectly classified. 

\begin{equation}
    MLC = H/L + \ln(\frac{R}{W})
\end{equation}

\begin{equation}
    MLC = H/L + \ln(\frac{R+0.5}{W-0.5})
\end{equation}

\begin{equation}
    MLC = H/L + \ln(\frac{R-0.5}{W+0.5})
\end{equation}

Table 3 summarizes the CAT information used for coding the process where CDI$_0$ is the initial starting estimate of ability (25th percentile).

\begin{table}[!h]
\caption{Summary of CAT Information}
\centering
\begin{tabular}{ |p{2cm}||p{3cm}|p{2cm}| }
\hline
Component & Parameters & Returns\\
\hline
Initializer & 25th percentile item & CDI$_0$ \\
\hline
Selector & Item bank, administered items, CDI$_t$,  & Index of next item\\
\hline
Estimator & item bank, CDI$_t$, administered items, response vector &  CDI$_{t+1}$\\
\hline
Stopper & administered items, SE$_M$ & True or False\\
\hline
\end{tabular}
\end{table}

Conventional and the novel MLC metric were timed in their computational efficiency while 'administering' cases and reported below. For completeness sake these were run on an Intel\textsuperscript{\tiny\textregistered} Core\textsuperscript{\tiny\texttrademark} i7-8565U CPU at 1.80GHz with 16GB RAM. 

\section{Results}
\subsection{Descriptive Results}
Distributions of the original CDI values for each dataset are shown in Fig. 4 and 5. In the frequency distributions, the data are shown in their 0.25$\theta$ width ‘bins’. The violin plots show the probability density functions of the data at each point by class. These plots indicate that the data are more separable in the Pulsar dataset compared to the MIMIC.

\begin{figure*}[!h]
\centering
\includegraphics[width=14cm]{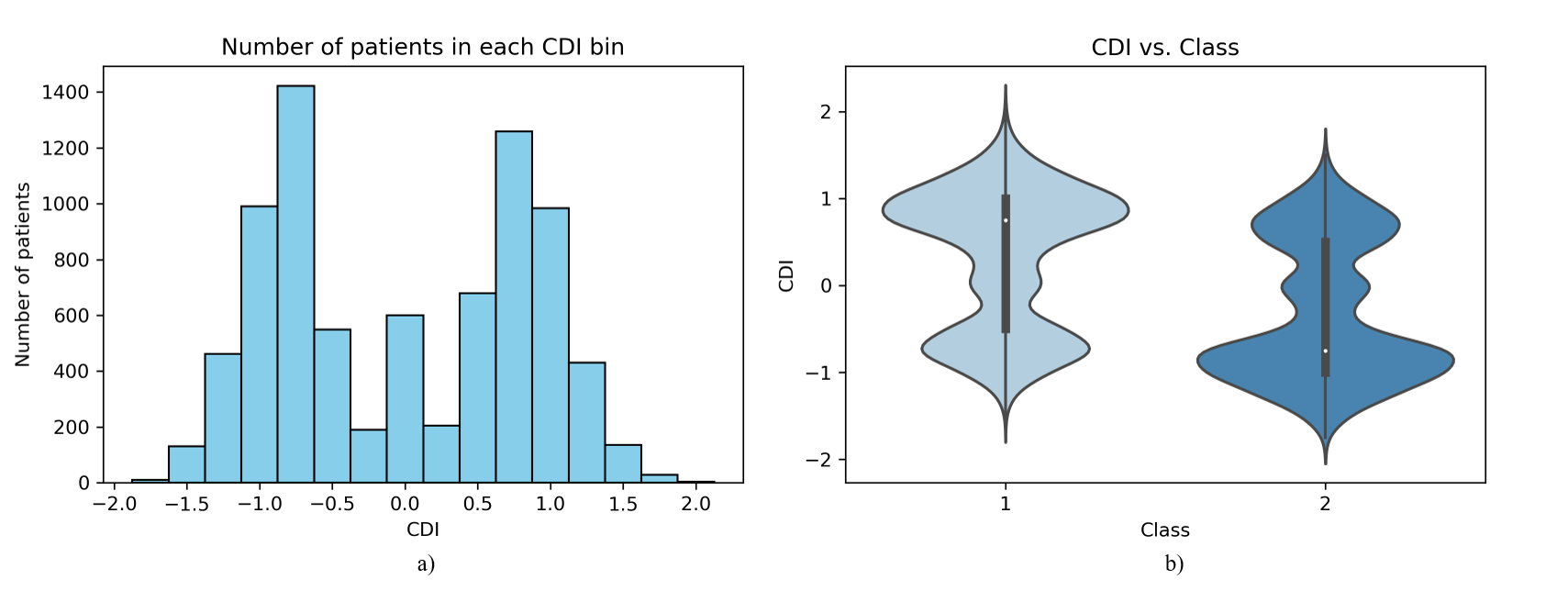}
\caption{MIMIC CDI distributions cumulatively (a) and by class (b)}
\end{figure*}

\begin{figure*}[!h]
\centering
\includegraphics[width=14cm]{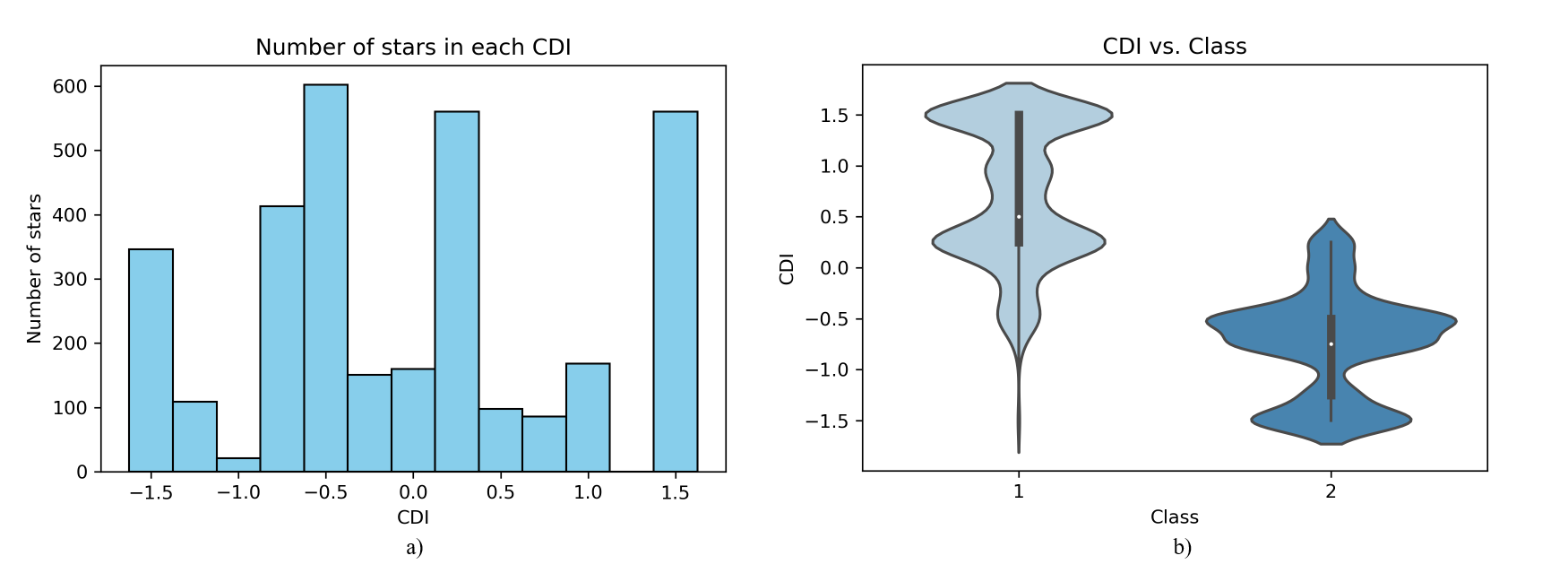}
\caption{Pulsar CDI distributions cumulatively (a) and by class (b)}
\end{figure*}

The CDI distributions for each class can be plotted against one another in a kernel density estimate (KDE) based on their actual class (1 versus 2) and the difficulty anticipated in assigning each case to its correct class. The class 1 data (‘dead’ in the MIMIC dataset and ‘pulsar star’ in the Pulsar dataset) were multiplied by -1 to ensure comparable CDI values across class. Using KDE plots, the distributions of these revised two datasets can be observed in Fig. 6. These provide an alternative visualization of the data that are presented in the violin plots. However, in the KDE plots the data for each class now have comparable ‘low’ and ‘high’ CDI values by virtue of having multiplied class 1 values by -1.

\begin{figure*}[!h]
\centering
\includegraphics[width=14cm]{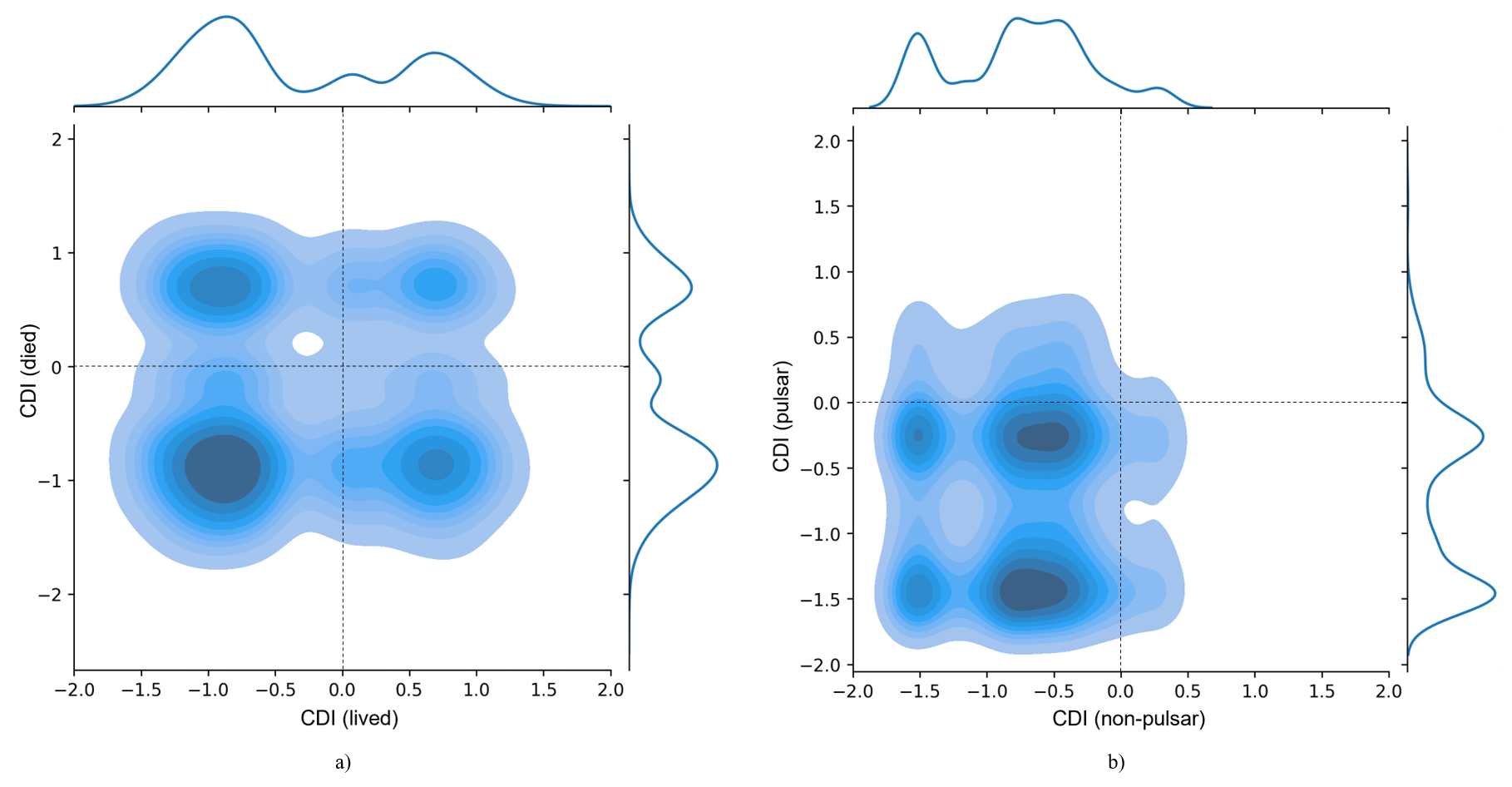}
\caption{KDE of difficulty contained within the set corresponding to each outcome class: a) MIMIC, b) Pulsar.}
\end{figure*}

\begin{figure*}
\centering
\includegraphics[width=14cm]{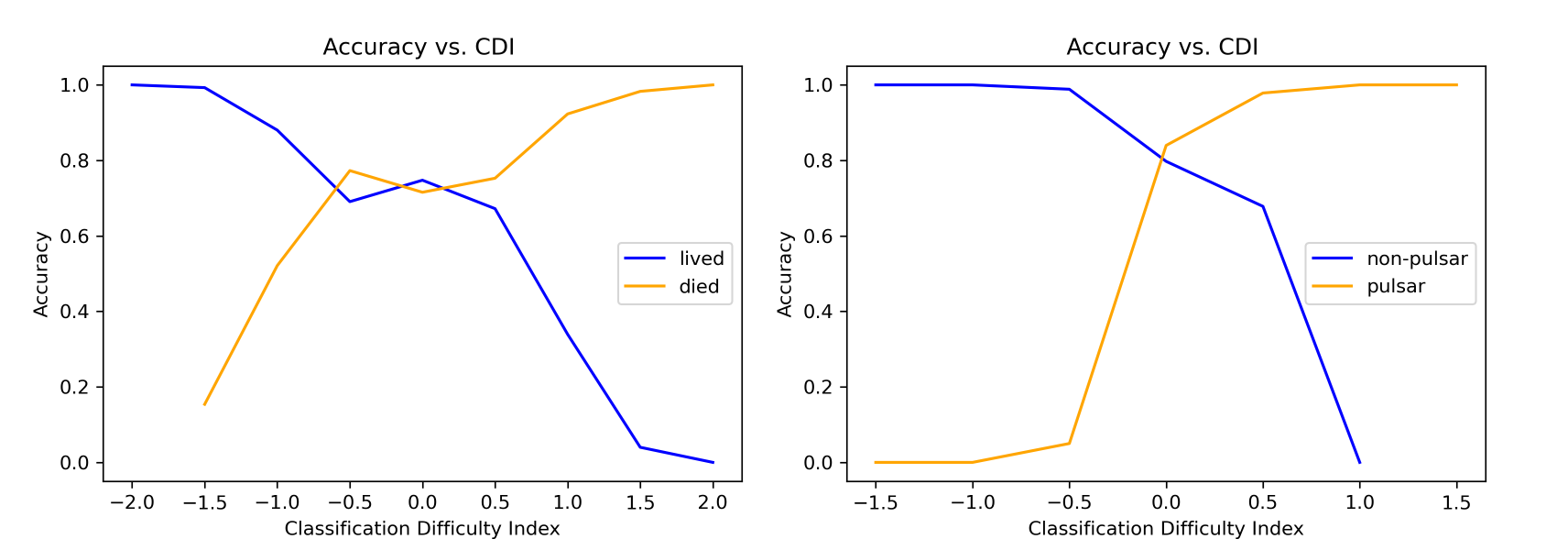}
\caption{Monotonicity of accuracy for MIMIC (a) and Pulsar (b) datasets}
\end{figure*}

\begin{table*}[!h]
\caption{MIMIC mortality - Evaluation Metrics}
\centering
\begin{tabular}{|c||c|c|c|c|}
\hline
Metric & Value & Data required (no.) & Comp. time (s) & Difficulty Adjusted\\
\hline 
Accuracy & 0.761 &  100\% (8078) & 0.520 & No \\
\hline
Precision & 0.756 &  100\% (8078) & 0.548 & No \\
\hline
Recall & 0.796  & 100\% (8078) & 0.389 & No\\
\hline
F1 & 0.752 & 100\% (8078) & 0.555 & No\\
\hline
AUC & 0.756 & 100\% (8078) & 0.66 & No\\
\hline
MLC (died) & 0.43  & 0.21\% (17) & 0.011\ & Yes\\
\hline
MLC (lived) & 0.78 & 0.22\% (19)  & 0.021 & Yes\\
\hline
\end{tabular}
\end{table*}

\begin{table*}[!h]
\caption{Pulsar star - Evaluation Metrics}
\centering
\begin{tabular}{|c||c|c|c|c|}
\hline
Metric & Value & Data required (no.)& Comp. time (s) & Difficulty Adjusted\\
\hline
Accuracy & 0.92 & 100\% (3278) & 0.34 & No\\
\hline
Precision & 0.967 &  100\% (3278)& 0.33 & No \\
\hline
Recall & 0.880  & 100\% (3278) & 0.29 & No\\
\hline
F1 & 0.922 & 100\% (3278) & 0.31 & No\\
\hline
AUC & 0.926 & 100\% (3278) & 0.32 & No\\
\hline
MLC (pulsar) & 0.12  & 0.55\% (18) & 0.019 & Yes\\
\hline
MLC (non-pulsar) & 0.32  &0.52\% (17)  & 0.013 & Yes\\
\hline
\end{tabular}
\end{table*}

\begin{figure*}[!h]
\centering
\includegraphics[width=14cm]{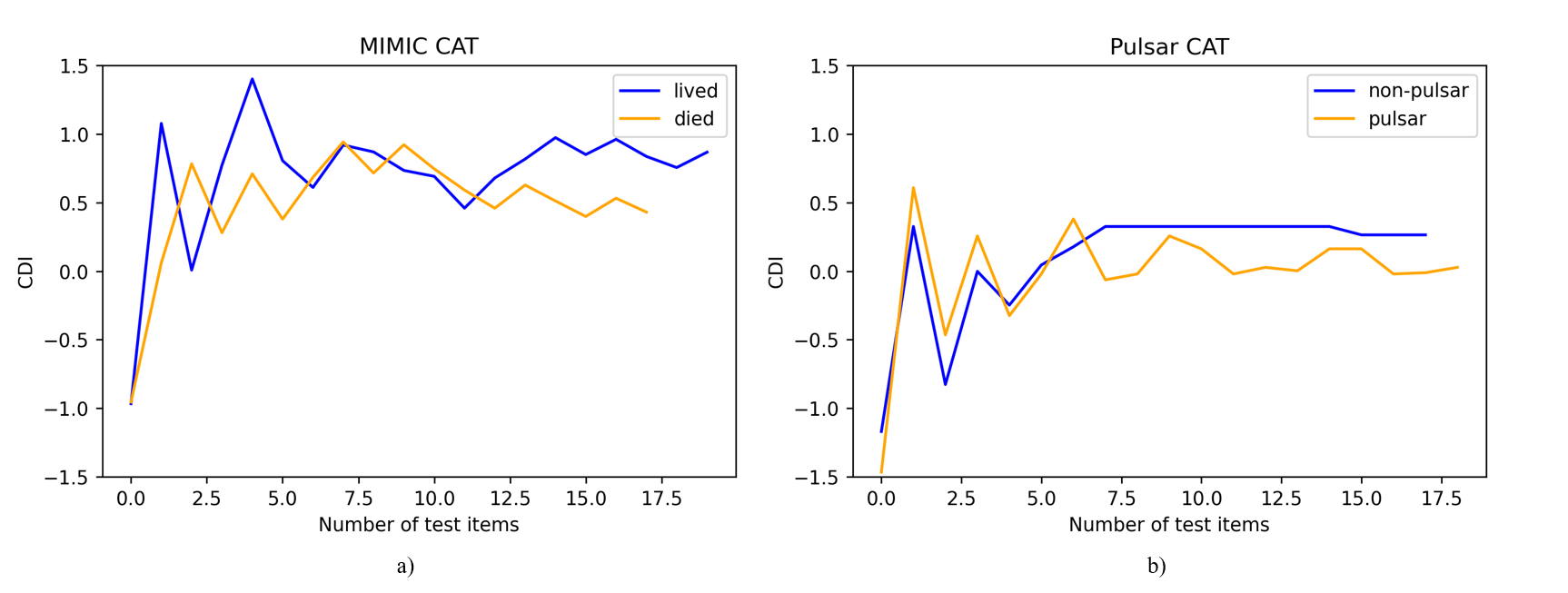}
\caption{Computer adaptive testing evaluation of each test}
\end{figure*}

It is noticeable from the KDE plots that the overall difficulty of the MIMIC data is higher relative to the Pulsar data. Most of the cases have CDI scores less than 0.0 in the Pulsar set, while many are much higher in the MIMIC set. This means that there are going to be more cases in the MIMIC data that are likely to be classified into the incorrect class relative to the Pulsar set.\\

\subsection{Machine Learning, Computer Adaptive Testing (CAT), and Machine Learning Capacity (MLC) Results}
The initialization estimation values for the CAT for each class in the MIMIC dataset was: -0.955 (died) and -0.967 (lived). For the pulsar dataset: -1.17 (non-pulsar) and -1.47 (pulsar). These values correspond the CDI 25th percentile value for each class tested. The 25th percentile was selected to provide a relatively 'easy' case for classification. Fig. 7 confirms the monotonicity of the relationship between CDI scores and classification accuracy. This is relevant insofar as it substantiates that CDI scores are a valid way to assess case difficulty. This figure was generated by using the traditional machine learning evaluation  metric of accuracy and plotting it against class (prior to inversion for CAT) and CDI.\\
Tables 4 and 5 show the traditional and MLC evaluation metrics on the two datasets. Of note, a column for computational time and whether or not difficulty was incorporated in the metric are additionally reported. The MIMIC data set shows solid classification with the model. Interestingly the MLC value for the cases that ‘died’ reached a CDI difficulty level of 0.43. This indicates the model was able to reliably correctly classify the ‘died’ cases with CDI scores almost a half standard deviation above the mean.  The MLC value for the cases that were ‘alive’ reached a CDI difficulty level of 0.78. This indicates the model was able to reliably correctly classify the ‘alive’ cases with CDI scores more than 3/4 of a standard deviation above the mean.\\
The Pulsar data, on the other hand, showed excellent classification. The MLC values indicate that the model was able to reliably correctly classify ‘pulsar’ cases with CDI scores just above the mean, and non-pulsar cases with CDI scores about 1/3 a standard deviation above the mean.  
It is also important to note the very small number of resources needed to ascertain the MLC values for all of the classes. Less than 1\% of the data were needed for these metrics to converge, and 5\% of the computing time of the traditional metrics.\\
Fig. 8 graphically represents how the MLC values are arrived at based on correct and incorrect responding in the CAT process. The level at which the MLC line asymptotes correspond to the level of case difficulty that can be reliably handled by the machine learning algorithm. Specifically, Fig. 8 b) in the 'non-pulsar' data hits a perfectly flat line as it has reached the maximum CDI contained within the Pulsar dataset for that class (0.34) as the ML algorithm is consistently getting these cases correct the selection process will continue to pose cases to the algorithm that are ideally more difficult but as our item selection is always limited by our CDI distribution in the dataset it will search the same vicinity until the $SE_M$ is reached. 

Fig. 9 shows how the summary MLC values for each class can be reported in an (x, y) coordinate system. The x-axis corresponds to the MLC value for class 1 and the y-axis corresponds to the MLC value for class 2. The MLC values for the MIMIC classes (0.78 - alive, 0.43 - dead) and for the Pulsar classes (0.32 – non-pulsar, 0.12 - pulsar) indicate the maximum points at which the machine learning algorithm is capable of differentiating each class. Summarizing the combination of the characteristics of the data set and the machine learning algorithm used in this manner provides insight into the relative robustness of these two regarding their capability in handling difficult to classify cases. While the Pulsar traditional measures suggested an excellent model, this was primarily due to the type of data with which it was presented. The MIMIC traditional measures suggested more modest classification accuracies. However, this model was capable of handling more difficult cases relative to the Pulsar model.\\

\begin{figure}[!h]
\centering
\includegraphics[width=8cm]{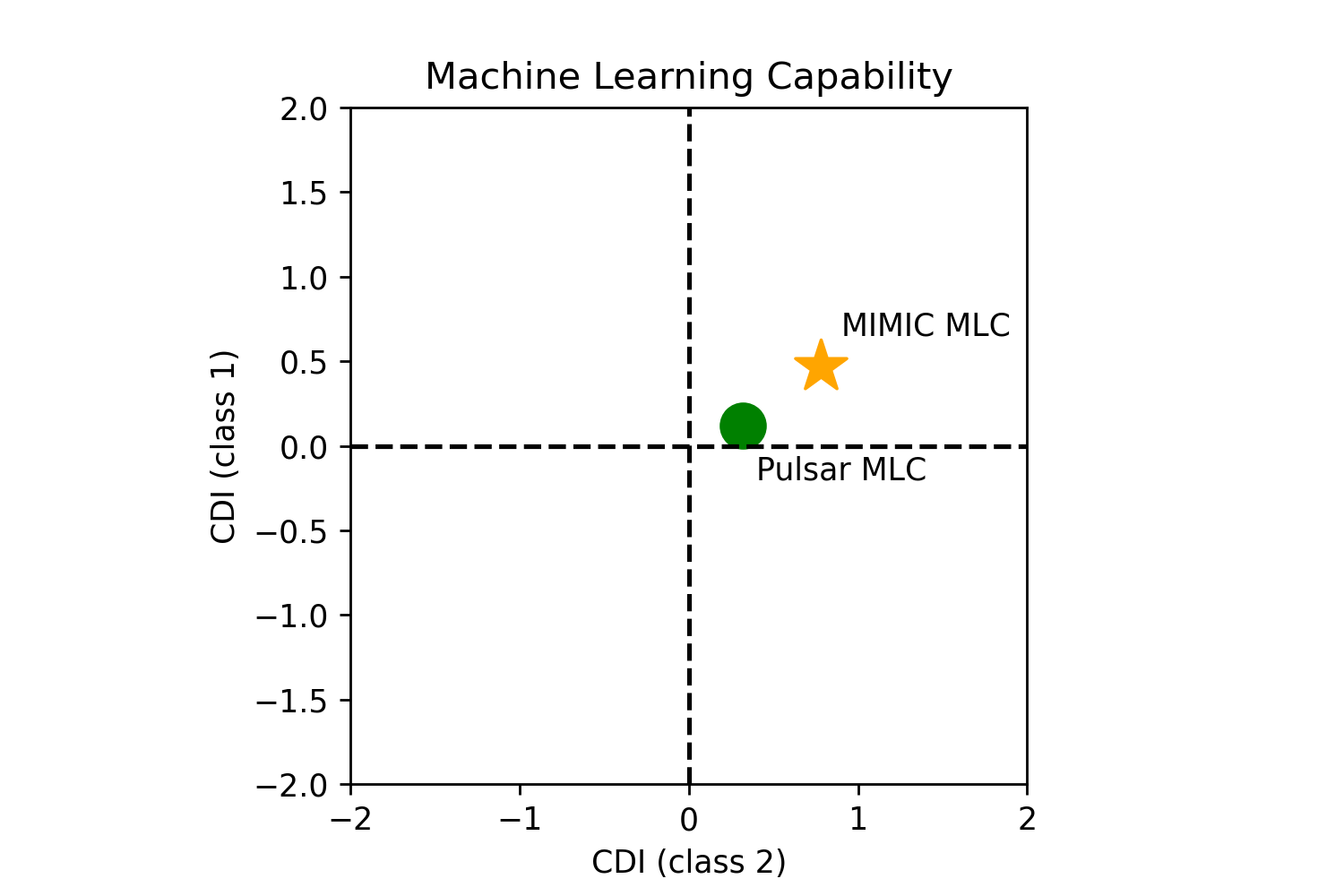}
\caption{Machine Learning Capability of Classifiers}

\end{figure}

\subsection{Practical Application of the Machine Learning Capability (MLC) Index}
These MLC values can now be used to assess whether or not a case should be classified based on the machine learning algorithm, or if the case characteristics exceed the limitations of the model. For example, assume a new patient comes into an intensive care unit (ICU) and their SAPSII data are inserted into the MIMIC-based machine learning algorithm generated in this research. The machine learning algorithm produces a class prediction (alive or dead) for that case. That patient’s SAPSII data will also be submitted to the IRT function with known feature parameters and thus a CDI score for the patient will be generated. \\
If the CDI score is within the maximum capability of the machine learning model, then we can be confident that the machine learning algorithm is able to classify that case.  This confidence is bolstered because the reliability of the stopping criterion for the MLC values was set to be very high (0.98). If the CDI score is outside the maximum capability of the machine learning model, then the case should have an intervention, with a field expert.\\  
Note that the CDI would need to be multiplied by -1 if the case was classified by the machine learning algorithm as ‘dead’ (class 1), as per the additive inversion that was necessary for the class 1 data CAT administration and the subsequent MLC metric. If the case was classified as ‘alive,’ there would be no need for the inversion before interpretation. 
Continuing with the example, assume that the case had a CDI score of 0.80 and was classified as ‘alive’. Since the classification is associated with class 2, we would directly assess the CDI against the class 2 MLC value of 0.78 and observe that it exceeds this value. Thus, we would assume that the case was too difficult to be classified by the algorithm and would need input from other sources (e.g., physician) to make the classification call. This is demonstrated in Fig. 10.\\
Working through a second example, assume that the case had a score of 0.80, but was classified as ‘dead’. Since this classification is associated with class 1, we need to invert the score to be -0.80 before interpretation. Since this score is below the MLC value of 0.43, we would assume that the machine learning algorithm was correct in making the classification that this case is highly likely to die.\\
Similarly with the Pulsar data: Assume a star’s integrated profile value CDI score was 0.10 and classified as a non-pulsar. Since this value is below the class 2 critical MLC value for a non-pulsar star, it can be assumed to be correctly classified as such. On the other hand if it was classified as a pulsar (and thus the CDI score inverted to be -0.10), this exceeds the class 1 value of -0.12. Thus, the astronomer might look for other evidence for correct classification of this star.\\
Testing the cases with CDIs below the MLC in either dataset for each class we get an accuracy of 94.7\% percent for pulsar star and 96.8\% for non-pulsar star. In the MIMIC dataset we achieve 91\% in the lived class and 87.5\% in the died class. 

\begin{figure*}[!h]
\centering
\includegraphics[width=14cm]{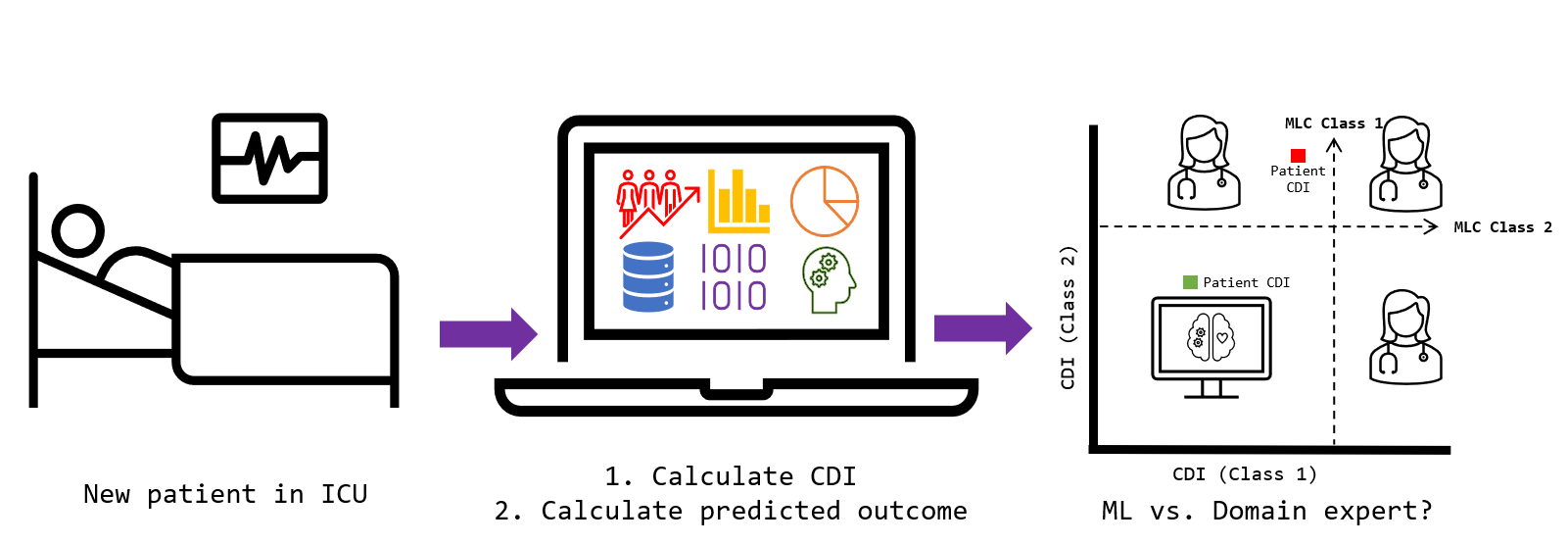}
\caption{Implementation of the MLC for real-world decision making}
\end{figure*}

\section{Discussion}
\subsection{General Findings}
The purpose of this research was to extend the use of CDI values generated from IRT models to assess the capability of a machine learning algorithm as it pertains to successful classification of difficult cases. This process led to the development and implementation of the MLC index. The MLC provides interesting and useful information to the end user.\\
First, it provides a summary index that integrates the performance of the machine learning algorithm with the uniqueness of the data set under investigation. This makes it possible to compare models as well as assess one’s own model. Second, it requires very little data - a benefit that many disciplines would find helpful particularly when users only have access to small datasets. Third, the CDI indices on which it is based are separate from the machine learning model. This avoids the circularity issue often run into where end-classification is used to assign weights to the features used in the algorithm. Fourth, although not usually a problem for many computer systems, the time taken to generate the MLC indices is much lower than for traditional machine learning evaluation metrics. 
The practical application possibilities of the MLC are profound. If one knows the CDI value of a case, and one knows the upper limits of the capability of a machine learning algorithm, it is possible to determine whether or not the case should be reliably classified by the algorithm. We can now add the MLC metric to our ML-based decision making repertoire to suggest the point at which humans need to intervene in case assessment. Up to the MLC, the practitioner can be confident that the algorithm will perform in a valid and reliable manner. \\
This approach demands a strong machine learning protocol be designed and precise CDI information for the case be ascertained. Use of such a methodology should be undertaken only with a solid understanding of the strengths and limitations of various machine learning models is imperative. The current and prior research suggests that the assessment of feature characteristics using IRT is a very valuable resource to set such CDI values as it provides robust estimates of the features’ characteristics when processed using a large number of cases \cite{Guilleux}. \\
Typically in a supervised machine learning problem data allocated to the 'test' category is anywhere between 10-30\% of the data. For which evaluations regarding specificity, sensitivity, recall, F1 and accuracy are calculated. The proposed method advocates for moving toward a single numeric per class outcome in supervised machine learning that takes into account case difficulty. Case difficulty is established independently of the classification performed by algorithms and as such remains an unbiased metric. As cases are normalized for difficulty in the dataset, the MLC represents both an ability metric in the more traditional sense of IRT literature and a cross-dataset metric.\\
Traditional machine learning model evaluation metrics are useful in that they provide an overall sense of the utility of a model on a data set. We are not advocating replacing them; instead we suggest that adding the MLC would provide complementary information to the investigator. This index provides a more nuanced understanding of which cases are difficult to classify and at what point they become so difficult that the machine learning algorithm requires additional input, such as a human expert. \\
As observed in the results, the Pulsar dataset was more separable and had a higher accuracy. But when bench-marked against the MIMIC dataset it was clear that this high accuracy was performing well due to the characteristics of the data set (i.e., specific cases and specific features), and not necessarily the capability of the machine learning model.\\
Using the MLC may be viewed as an evolving process when the CDI is not a static measure of a case's difficulty. This would occur when the predictor values are in flux (e.g., changing patient data over time). In such instances the CDI acts a temporally responsive metric and may change in a direction that demands thoughtful application for the continued use (if the CDI is below the MLC of model) or disuse (if the CDI is above the MLC of model) of the machine learning model as it pertains to a specific case at a given point in time. Following this logic, researchers and end users are encouraged to perform ad-hoc investigations into which features and their respective values are driving a given case to have a CDI that may exceed the limitations of the ML model. With this insight, attempts can be made to correct or target these predictor variables (if modifiable) with the goal of moving the case into an easier-to-classify state that would be within the capability of the MLC of the trained model. This easier classification state should be beneficial to both ML-based and human expert decision-making paradigms.\\
An understanding of when and why algorithms should be used to support human decision-making will facilitate their uptake and allow for the judicious expenditure of labor-intensive human resources. However, many ethical and legal issues surround real-world deployment of ML algorithms. Some of these include transparency, known or unknown biases, social benevolence, and privacy of information \cite{Vollmer}. Knowing a-priori that an individual case is well encapsulated within the limitations of the model is important to build trust with domain experts as well as address the issues that may arise in answer to the question ‘can/should this be handled by our ML model?’ The MLC metric devised here can address some of these concerns as the CDIs on which it is based are calculated independent of the actual ML models using analytics that are transparent, and any biases against race, gender, etc. can be assessed by way of IRT multi-group analyses. These possibilities are important to explore based on extant literature that suggests expected differences \cite{Vollmer}. Decisions are becoming more complex with the addition of new information regarding diagnostic tools, treatment options, possible side effects, human burnout, and cost effectiveness \cite{CIHR}. Maximizing ML and human resources while simultaneously engendering transparent insight should be a priority for industry and government stakeholders and the MLC framework’s utility as a decision-support system has wide-ranging implications.

\subsection{Future Research}
Several areas of future research suggest themselves from this study. We used a single mode of data (i.e., numeric) to showcase the MLC. However, incorporating multi-modal data into machine learning algorithms and IRT models would be an important next step in this research. The data in our sets had only two classification outcomes. We hypothesize that the MLC index process would extend well to multi-class outcome data. In addition, we used balanced data sets. With an imbalanced dataset, depending on the class that is imbalanced and what CDI values are contained within the additional cases, there may exist a higher MLC for that class. This would be particularly true if these case CDI values are close to zero or higher. This is an empirical question that can be answered in future research.\\
Though this process was demonstrated using neural networks, this method can be extended for use in all supervised machine learning models as well as multi-class outcomes.\\
The MLC index approach is anticipated to extend to areas of supervised machine learning where there is a need to determine the maximum capability of the algorithm, and where other knowledge (e.g., human) needs to be drawn upon. This is likely to be of interest in many fields where much effort has been focused and algorithms are compared to their human counterparts. Determining where the machine-human dividing line is located has, to this point, remained elusive. The MLC provides a step forward in attempting to solve this problem. A critical follow up to this research would be to trial this approach in a real-world setting and assess its anticipated performance.

\section{Conclusion}
The MLC index represents a reliable and novel way to evaluate and report in a standardized manner the classification performance of a supervised learning algorithm for each class of outcome. It is less reliant on the total number of correct versus incorrect classifications, and instead focuses on which ones are correct and which ones are incorrect. This measure uses far fewer data points, and is computationally more efficient. While a neural network was used as the example of traditional machine learning approach in the current research, the process is suitable for use in all supervised machine learning models and can also be extended to multi-class outcomes. The MLC index provides a metric by which the limitations of machine learning models can be ascertained, and thus at what point other interventions need to override the machine learning algorithm.


%


\ifCLASSOPTIONcompsoc
  \section*{Acknowledgments}
\else
  \section*{Acknowledgment}
\fi

The authors would like to thank MIT for the MIMIC database, the UCI repository for the Pulsar dataset utilized in this study and Dr. Theresa Kline for reviewing earlier versions of the manuscript.

\ifCLASSOPTIONcaptionsoff
  \newpage
\fi



%

%

\begin{IEEEbiography}[{\includegraphics[width=1in,height=1.25in,clip,keepaspectratio]{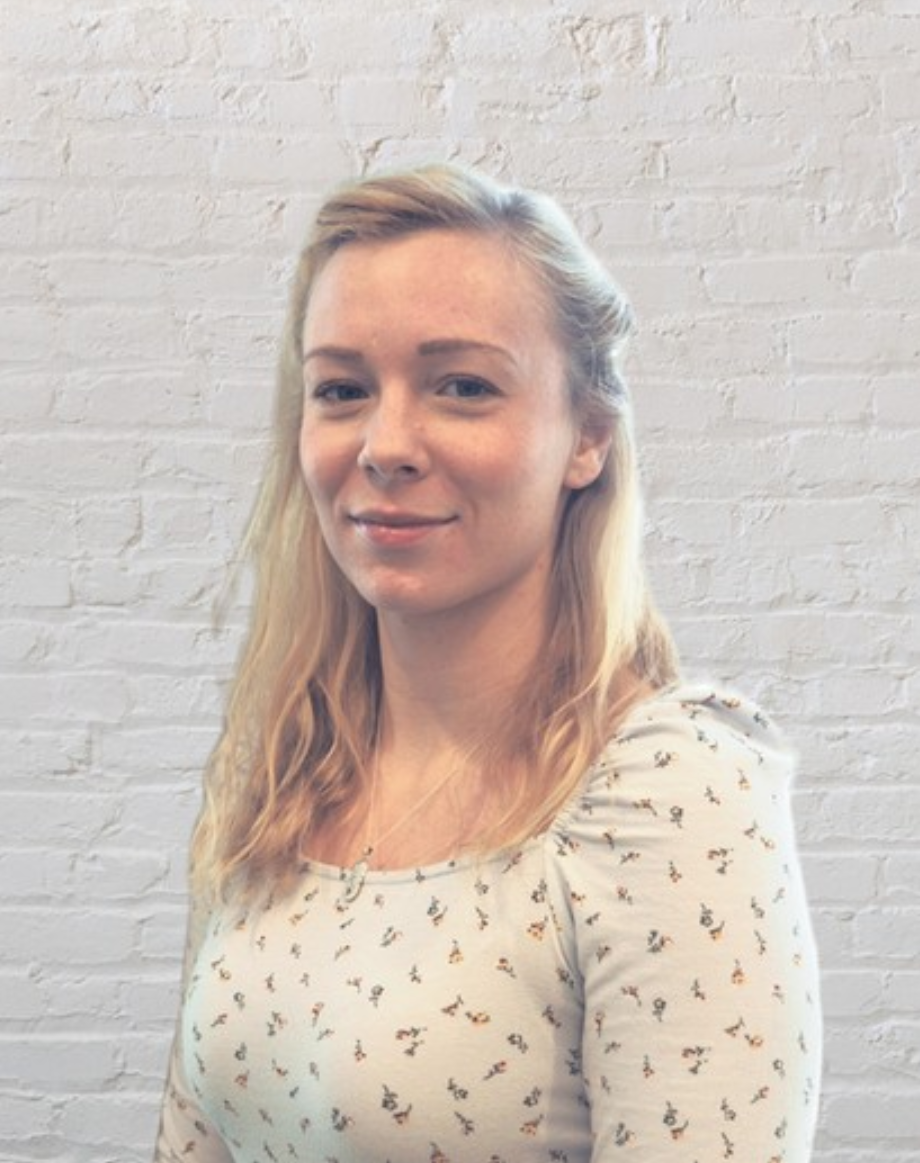}}]{Adrienne Kline}
Dr. Kline completed her PhD in Biomedical Engineering and MD at the University of Calgary. Prior to this she completed her BSc in electrical engineering with a specialization in biomedical engineering from Indiana Institute of Technology. While completing the medical portion of her training, she has been a research assistant in the Data Intelligence for Health (DIH) Lab, performing independent research.\\
She is currently pursuing a postdoctoral fellowship at Northwestern University.\\
Her primary research interests lay in creating novel methodologies that better enable real-world applications of machine learning, particularly in the health sector. Her research applies to artificial intelligence, data science, machine learning, biostatistics and natural language processing.\\
She has won multiple national academic accolades, including Natural Sciences and Engineering Research Council of Canada (NSERC) and Canadian Institutes of Health Research (CIHR) awards for her work. 
\end{IEEEbiography}

\begin{IEEEbiographynophoto}{Joon Lee}
Dr. Joon Lee is the Director of the Data Intelligence for Health Lab and an Associate Professor of Health Data Science in the Departments of Community Health Sciences and Cardiac Sciences, Cumming School of Medicine, University of Calgary. He is a member of the Libin Cardiovascular Institute of Alberta and the O'Brien Institute for Public Health.\\
He holds a PhD in Biomedical Engineering from the University of Toronto, and a BASc in Electrical Engineering from the University of Waterloo. He also completed a postdoctoral fellowship in Medical Data Science at the Harvard-MIT Division of Health Sciences and Technology. Prior to joining the University of Calgary, he held a faculty appointment in the School of Public Health and Health Systems, Faculty of Applied Health Sciences, University of Waterloo, for 6 years.\\
His primary research interest is in transforming health data from various sources into useful information and knowledge. His research applies data science, machine learning, artificial intelligence, natural language processing, mobile technology, and biostatistics to several health fields including intensive care medicine, aging, and population health surveillance.\\
In 2016, he was a recipient of the Early Researcher Award from the Ontario Ministry of Research, Innovation and Science.
\end{IEEEbiographynophoto}





\end{document}